\documentclass[letterpaper, 10 pt, conference]{ieeeconf}  

\IEEEoverridecommandlockouts                              

\usepackage{graphics} 
\usepackage[pdftex]{graphicx}
\graphicspath{{./Figures/,./}}
\DeclareGraphicsExtensions{.pdf,.jpeg,.png,.eps}
\usepackage{epsfig} 
\usepackage{amsmath} 
\usepackage{amssymb}  
\usepackage[export]{adjustbox}
\usepackage[]{subfig}
\usepackage{lipsum}
\usepackage{tabularx,booktabs}
\usepackage{textcomp}
\usepackage{algorithmic}
\renewcommand{\baselinestretch}{0.97}

\usepackage{cite}
\usepackage{cuted}
\usepackage{tikz}
\usepackage{float}
\usepackage{bm}
\usepackage{algorithm}

\newcommand{\vg}[1]{\bm{#1}}
\renewcommand{\v}[1]{\mathbf{#1}}
\newtheorem{remark}{Remark}

\usepackage{color}

\usepackage[font=footnotesize]{caption}

\makeatletter
\newcommand\mypicture[2][\textwidth]{%
  \setbox0\hbox{\includegraphics[width=\textwidth]{#2}}
  \ifdim\ht0>\dimexpr\pagegoal-\pagetotal\relax
    \@latex@warning{This picture might be oddly placed}\fi
  \begin{strip}
    \includegraphics[width=#1]{#2}
  \end{strip}
}
\makeatother

\title{\LARGE \bf
Simultaneous Trajectory Optimization and Force Control with Soft Contact Mechanics
}
\author{\large Lasitha Wijayarathne$^{1}$, Qie Sima$^{2}$, Ziyi Zhou$^{2}$, Ye Zhao$^{2*}$ and  Frank L. Hammond III$^{1*}$, \textit{IEEE Member}
\thanks{$^{1}$Lasitha Wijayarathne and Frank L. Hammond III are with the Woodruff School of Mechanical Engineering and the Coulter Department of Biomedical Engineering at the Georgia Institute of Technology. {\tt\small frank.hammond@me.gatech.edu.}}%
\thanks{$^{2}$Qie Sima, Ziyi Zhou, and Ye Zhao are with The Laboratory of Intelligent Decision and Autonomous Robots at Woodruff School of Mechanical Engineering, Georgia Institute of Technology. {\tt\small ye.zhao@me.gatech.edu.} (*co-corresponding authors)}
}

\begin{document}
\maketitle 

\thispagestyle{empty}
\pagestyle{empty}
\begin{abstract}
Force modulation of robotic manipulators has been extensively studied for several decades but is not yet commonly used in safety-critical applications due to a lack of accurate interaction contact modeling and weak performance guarantees - a large proportion of them concerning the modulation of interaction forces. This study presents a high-level framework for simultaneous trajectory optimization and force control of the interaction between manipulator and soft environments. Sliding friction and normal contact force are taken into account. The dynamics of the soft contact model and the manipulator dynamics are simultaneously incorporated in a trajectory optimizer to generate desired motion and force profiles. A constrained optimization framework based on Differential Dynamic Programming and Alternative Direction Method of Multipliers has been employed to generate optimal control inputs and high-dimensional state trajectories. Experimental validation of the model performance is conducted on a soft substrate with known material properties using a Cartesian space force control mode. Results show a comparison of ground truth and predicted model based contact force states for multiple Cartesian motions and the validity range of the friction model. The proposed high-level planning has the potential to be leveraged for medical tasks involving manipulation of compliant, delicate, and deformable tissues.  
\end{abstract}


\section{Introduction}
Robotic applications in the medical domain have gained increasing attention over the past few decades \cite{Solis2016,Shademan2016}. Within medical domain, planning and control of the interaction forces between a robot and its environment are essential to a variety of safety-critical tasks. For instance, the interaction force should be modulated accurately in compliant environments, such as those in a surgical setting, micro-assembly, or biological tissue manipulation. Force control based on identifiable physical models is essential to identify instability modes (e.g., those caused by the bandwidth and structure) and maintain reliable force interaction to guarantee safety. 
Thus, a model-based trajectory planning method with a high-fidelity contact force model is essential for successful deployment with satisfactory motion tracking and contact force performance. \par
\begin{figure}[t!] 
\centering
\includegraphics[width=1\linewidth]{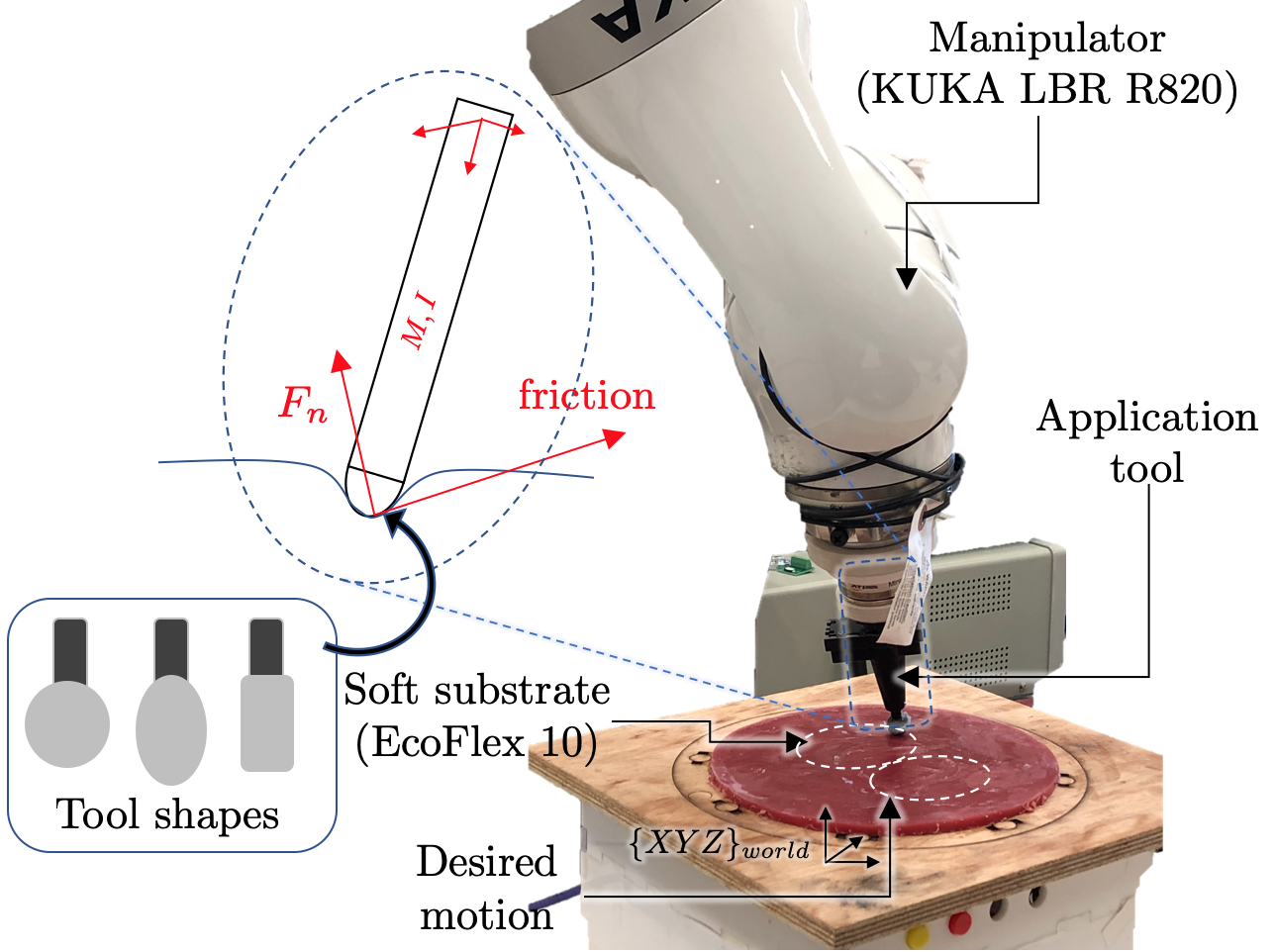}
\caption{KUKA manipulator experimentation setup. A manipulator performing a force controlled motion task on a soft surface. $XYZ_{\{world\}}$ is the world frame.}
\label{fig:setup_intro}
\vspace{-15pt}
\end{figure}

Unlike rigid contact models, soft contact models are subject to challenges posed by non-linear material properties and non-uniformity as well as intensive computation burden. Numerous contact models have been presented in the literature to model interactions involving elastic deformation  \cite{elandt2019pressure}. These models have broad applications and are essential in many engineering areas such as machine design, robotics, multi-body analysis, to name a few. For contact problems that involve elasticity, Hertz adhesive contact theory has been well established\cite{Johnson1985ContactMechanics}. In this study, we focus on robotic tasks interacting with soft tissues, the contact behavior of which is determined by not only external and viscous forces, contact geometry, but also material properties (see Figure \ref{fig:setup_intro}). The soft contact mechanics are crucial in the physical model identification for applications in surgical robots.
\par
Simultaneous trajectory generation and force control enable sophisticated manipulation tasks while interacting with complex objects. As a promising approach along this direction, trajectory optimization with contact models has been extensively investigated in the robotics community \cite{posa2014direct, mordatch2012discovery, manchester2020variational, sleiman2019contact,marcucci2016two, neunert2017trajectory, neunert2018whole}. By incorporating the contact dynamics into the optimization, contact-dynamics-consistent motions can be planned for complex robot behaviors, such as dynamic locomotion or dexterous object manipulation. A majority of them focused on rigid contact dynamics \cite{posa2014direct, mordatch2012discovery, manchester2020variational, sleiman2019contact}; whereas \cite{ marcucci2016two, neunert2017trajectory, neunert2018whole} directly integrated a soft contact model inside the system dynamics, and implicitly optimized both contact force and other control inputs. In \cite{fahmi2020stance}, a soft contact model was taken into account inside the optimization formulation for Whole-Body locomotion Control. However, most of the works above assumed spring-damper type soft contact models, which still largely mismatched the contact surface deformation or elasticity in reality. In this study, we leverage a distributed optimization algorithm proposed recently in \cite{ZhouAcceleratedDynamics} to solve a constrained trajectory optimization with a high-fidelity soft contact model. This algorithm provides a general optimization framework to iteratively solve decomposed rigid body dynamics, articulated robot kinematics, and inequality constraints in a computationally efficient manner.

\par 
Manipulator contact models are naturally framed and executed in the task space. In safety-critical tasks such as soft material manipulation and medical applications, force-torque control plays a significant role. These include interaction with humans in proximity or with direct physical contact. In approaching the contact interaction problem, data-driven  techniques have been explored to learn the interaction between robotic manipulators and the environment \cite{Pereida2018Data-EfficientTransfer, ParigiPolverini2019MixedApproach}.
Unlike rigid contacts, the soft environment is stochastic in nature. Thus, it is challenging to learn the contact model and robot dynamics simultaneously through data. In this study, we present a model for contact interaction and embed it into the high-level motion planning via enforced constraints.
 \par
\noindent The main contributions of this study are listed below:
\begin{itemize}
\item Presentation of a dynamic interaction model based on soft contact mechanics for a predefined geometry with Hertz visco-static theory.
\item Incorporation of the interaction model with a distributed DDP-type trajectory optimization with constraints to generate the desired Cartesian path and force profile.
\item Experimental validation of the derived contact dynamical model and implementation of the proposed trajectory optimization algorithm.
\end{itemize}

\section{Related Work}
\label{sec:related work}
Elastic contact mechanics \cite{Johnson1985ContactMechanics} have been extensively studied in various research fields where contact modeling is imperative for safety and performance requirements. Existing works in \cite{Gilardi2002LiteratureModelling, Schindeler2018OnlineLinearization,Lefebvre2003EstimatingMotion,Marhefka1999ASystems} have used soft contact models for both modelling and control. These works include quasi-static assumptions and studies of  \cite{Pappalardo2016Hunt-CrossleySurgery, Sun2018InternationalDamping} explore cases where high-velocity impacts on soft material are considered. In the impact cases, visco-elastic models have been widely investigated. For instance, studies in \cite{Pappalardo2016Hunt-CrossleySurgery,Gilardi2002LiteratureModelling} compared various visco-elastic models with experimental validations. Overall, a majority of these works have shown that the Hertzian-based Hunt-Crossey model is the one most suitable for visco-elastic cases.
Furthermore, fundamentals of frictional sliding motion are established in the works of  \cite{Howe1996PracticalManipulation, mason1986mechanics}, where the main focus is on rigid body contacts but generalizable to soft contacts. More recent works in \cite{Mu2019RoboticMotion, Fazeli2017ParameterContact} propose contact-area-based models. 

\par

Trajectory optimization (TO) is a powerful tool to generate reliable and intelligent robot motions. Various numerical methods have been proposed to solve a TO \cite{betts1998survey, jacobson1970differential, betts2010practical}. Among them, Differential Dynamic Programming (DDP) and iterative Linear Quadratic Regulator (iLQR) have aroused much attention in solving TO in the context of unconstrained problems, where only dynamics constraint is enforced in the forward-pass. The Ricatti-like backward pass in DDP or iLQR effectively reduces the complexity of solving an approximated LQR problem over the entire horizon, and the optimization is solved in an iterative fashion. In \cite{TassaSynthesisOptimization}, DDP is used in a balancing task of a humanoid robot with high degrees of freedom (DoFs). A more recent work \cite{Koenemann2015Whole-bodyHumanoid}, demonstrates a Model Predictive Control (MPC) implementation based on DDP. However, standard DDP algorithms are not capable of addressing constraints. In \cite{TassaControl-LimitedProgramming, Xie2017DifferentialConstraints, howell2019altro}, DDP-type variants are proposed to cope with state and control constraints. Instead, our approach employs an augmented Lagrangian method named as Alternating Direction Methods of Multipliers (ADMM) \cite{o2013splitting, o2016conic, sindhwani2017sequential, ZhouAcceleratedDynamics} to address various constraints. This ADMM framework is capable of tackling more constraints by introducing additional optimization blocks, making the algorithm suitable for parallel computing.

\par

\section{Soft contact modeling}
\subsection{Contact modeling via Hertz's theory}
\label{contact interaction model}
\begin{figure}[t]
\centering
\includegraphics[width=.8\linewidth]{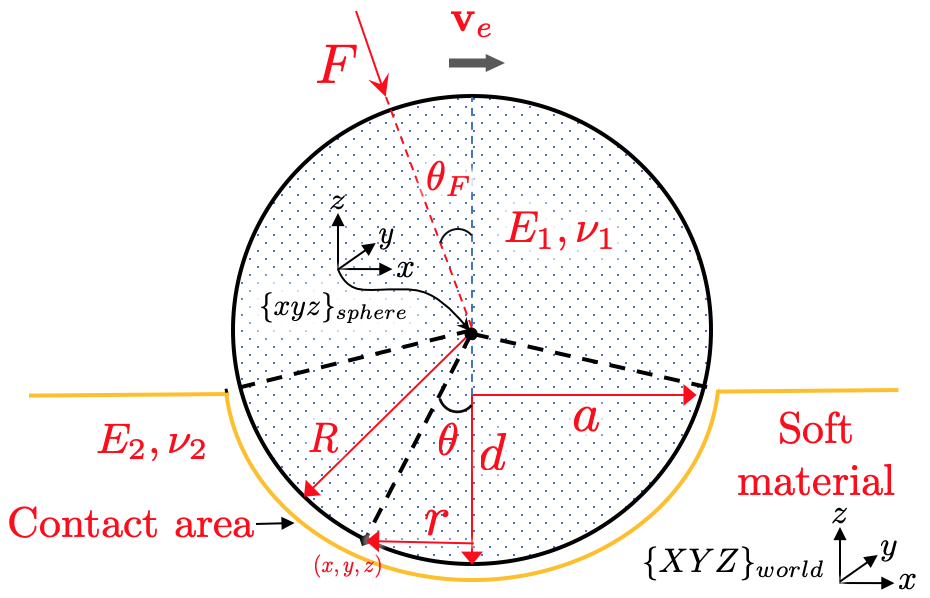}
\caption{A graphical illustration of the soft contact model between surface and the end-effector tool}
\label{fig:contact_ball}
\vspace{-15pt}
\end{figure}

In this section, we model the interaction dynamics between an application tool mounted on a manipulator and a soft tissue in terms of contact geometry and mechanics.
In the example shown in this study, the contact part of manipulation is assumed to be a
spherical indentation (for simplicity, but not limited to). Further, we assume that the application tool used is rigid and has a high stiffness compared to the contact surface. Along with these assumptions, we derive a dynamic model based on the contact friction theory and Hertz visco-static theory. According to Hertz's theory, the largest static indentation is achieved at the central point of the circle (see Figure~\ref{fig:contact_ball}) and can be expressed as:
 \begin{equation}
     d=\left[\frac{9F^{2}}{16E^{2}R}\right]^{\frac{1}{3}}
     \label{eq:quasi_static_model}
 \end{equation}
where $E$ is the reduced Young's modulus of tool and surface, $R$ is the radius of the tool end,  $F$ is the force imparted on the surface by manipulator end-effector. Combined Young's modulus of the tool and the soft contact surface material can be lumped to one term as:
 \begin{equation*}
     \frac{1}{E}=\frac{1-\nu_1^2}{E_1}+\frac{1-\nu_2^2}{E_2}
 \end{equation*}
where $E_1$, $E_2$ and $\nu_1$, $\nu_2$ are Young's moduli and Poisson ratios of the end-effector and contact surface material, respectively.
In our scenario, we assume the contact part as a rigid object and thus the Young's module of the spherical cap $E_2$ is approximated as infinity. Accordingly, we have $E=E_1/(1-\nu_1^2)$.
The deformation and stress distributions on the surface are approximated by the universal Hooke's law and Hertz's theory. Details of normal, radical, and hoop (i.e., moving direction) stress distributions within the contact area in the cylindrical coordinate system are provided in the Appendix.



Accordingly, the deformation distribution is derived from the stress distribution equations as follows:
\begin{equation*}
u_z=\begin{cases}
  \frac{3\pi}{8a}\Big[\frac{1-\nu^2}{E}\Big]p_m(2a^2-r^2), & (r\leq a) \\ \\
  \frac{3}{4a}\Big[\frac{1-\nu^2}{E}\Big]p_m\Big[(2a^2-r^2)\sin^{-1}\big(\frac{a}{r}\big) \\ \quad + a(r^2-a^2)^{\frac{1}{2}}\Big], & (r \geq a)
    \end{cases}
\end{equation*}
where $p_m=F/(\pi a^2)$ is the average stress applied in contact part by manipulation and $a=\sqrt{Rd}$ is the radius of contact area (see Figure~\ref{fig:contact_ball}).
The dynamic contact model for a contact spherical cap (i.e., spherical geometry) is applied with a force vector $F$ at an angle $\theta_F$ to the perpendicular and moves in a circular path of radius $R$ with a uniform velocity $\mathbf{v}_e$ in frame $\{sphere\}$. It represents the scenario of manipulating an application tool to work with soft tissues.
For simplicity, our model focuses on sliding friction and ignores other frictional sources such as adhesion and rolling induced by deformation.
Due to the symmetry of our contact scenario, $\sigma_\theta$ represents the principal stress within the contact circle. Thus, we can represent the stress tensor of any contact point $(r,\theta,z)$ in cylindrical coordinates relative to frame $\{sphere\}$ via the Cauchy stress theory \cite{Johnson1985ContactMechanics}.
\begin{equation}
  \vg \sigma= {\left[ \begin{array}{ccc}
\sigma_r & 0 & \sigma_{rz}\\
0 & \sigma_\theta & 0\\
\sigma_{zr} & 0 & \sigma_z
\end{array} 
\right ]}
\end{equation}
Since the task is defined in the Cartesian frame, we convert parameters to Cartesian coordinates from cylindrical coordinates. The stress tensor in Cartesian coordinate is $\vg \sigma_c= T^T \vg \sigma T$, where the transformation matrix $T$ is defined in the Appendix \ref{sec:appendix}.

At an arbitrary point on contact surface $(x, y, z)_{\{sphere\}}$, the normal vector from this point to centroid of spherical cap is $\mathbf{n}=[s\theta \quad 0 \quad c\theta]^T$.\footnote{we denote $\cos \theta = c \theta$ and $\sin \theta = s\theta$.} Then, the normal stress of the contact surface is $\sigma_n=\mathbf{n}^T\vg \sigma_c \mathbf{n}$ with
\begin{equation*}
\sigma_n = \sigma_r c^2\theta s^2\theta + \sigma_\theta s^4\theta+ \sigma_z c^2\theta+2 \sigma_z s\theta c^2\theta
\end{equation*}
Given this stress expression, the overall friction force of the contact surface is represented as
\begin{align}
    df=\mu \sigma_n dS=\mu \sigma_n\times 2\pi r \frac{dr}{c\theta} \\
    f=\int dfc\theta= 2\pi \mu \int_{0}^{a} \sigma_n rdr
    \label{eq. friction_integral}
\end{align}
where, $df, dr, dS $ are the differential elements of the friction, $r$ and contact area. In the surface normal direction, it is assumed that the surface is in contact with the end point of the tool.  As a result, Eq. ~(\ref{eq:quasi_static_model}) always holds. The derivative form of Eq.~ (\ref{eq:quasi_static_model}) is
\begin{equation}
    \dot{z}=-\dot{d}=-\left[\frac{1}{6E^{2}RF_z}\right]^{\frac{1}{3}}\dot{F_z}
    \label{eq:quasi-static-dir}
\end{equation}
where $z$ represents the position along the surface normal direction of the contact point and force along the normal direction is defined as $F_z = F\cos{\theta_F}$.
In the moving direction, by using Eq. ~(\ref{eq. friction_integral}) and $\mu F_z$ to get the frictional force caused by the normal force $F_z$, total friction is:
\begin{equation}
    F_\theta=f=\mu F_z\left[1+(2\nu-1)\frac{3a^2}{10R^2}\right] + k_d\v v_e
    \label{eq:friction_model}
\end{equation}
where $k_d$ is a damping coefficient in the moving direction. By substituting $a=\sqrt{Rd}$ and Eq.~(\ref{eq:friction_model}), we have the derivative form of Eq.~(\ref{eq:friction_model})
\begin{equation}
    \label{eq:friction_dyn}
    \dot{F_\theta}=\dot{f}=\mu \dot{F_z}+\frac{3 \mu (2\nu-1)}{10R}\left(\dot{F_z}d-F_z\dot{z}\right) + k_d\dot{\v v}_e
\end{equation}
In the radial direction, we have $F_r=mv^2/R_c$, where $m$ is the mass of the tool and $R_c$ is the radius of the curvature path. Then the derivative of $F_r$ is
\begin{equation}
    \label{eq:friction_dyn_orthogonal}
   \dot{ F_r}=2m v\dot{v}/R_c
\end{equation}

\subsection{Other visco-elasticity contact models}
Apart from the Hertz model that we used, there exist other visco-elastic contact models \cite{Pappalardo2016Hunt-CrossleySurgery,Gilardi2002LiteratureModelling} that are proven to effectively estimate the interaction between the soft material and tools. The well-received models are:
\subsubsection{Kelvin–Voigt (KV) Model}
model consists of a spring in parallel with a damper. 
\begin{equation*}
    F_e = K{\delta x} + \underbrace{D\dot{\delta x}}_{A_1}
    \vspace{-5pt}
\end{equation*}
\subsubsection{Hunt and Crossey Model}
 This model is a modification to the Hertz quasi-static model with a non-linear damping term. 
 \begin{equation*}
    F_e = K{\delta x}^n + \underbrace{\lambda{\delta x}\dot{\delta x^n}}_{A_2}
    \vspace{-3pt}
 \end{equation*}
where, $\delta x$ is the indentation depth, $K$ and $D$ are stiffness and damping coefficients,  and $F_e$ is the contact force associated with it along the surface normal.
 \par
Visco-elastic models yield better results when there exist hard impacts or high velocities involved in the direction of penetration. This is due the presence of $A_1$ and $A_2$ damping terms in the model.  The work of \cite{Pappalardo2016Hunt-CrossleySurgery} presented a quantitative study on a comparison of different visco-elastic models. This is in-fact useful in force modulation in non-stationary environments where penetration depth changes frequently. We limit our scope to the Hertz quasi-static model in our formulation, but it could be extended to the models mentioned above.
\begin{figure}[t] 
\centering
\includegraphics[width=\linewidth]{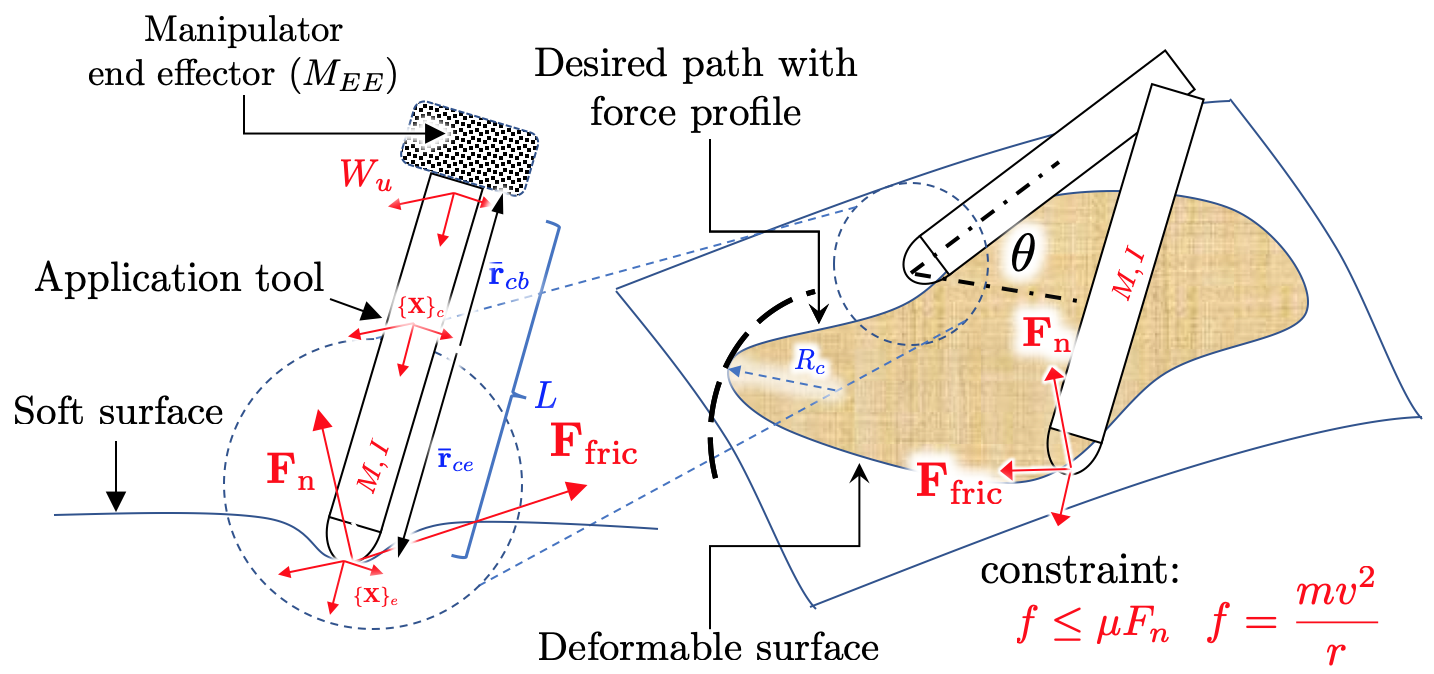}
\caption{Simultaneous motion and force control on a soft surface. Mathematical notations are shown for the contact forces and deformable surface. }
\label{fig:modulation_setup}
\vspace{-18pt}
\end{figure}
\label{sec:manipulator model}

\section{Manipulation dynamics and tool model}

\subsection{Manipulator and application tool model}\label{manipulator and tool models section}
This section formulates the manipulator dynamics as well as the dynamics of the tool in the optimization.
We use the wrench control capability of the manipulator as described in \ref{manipulator control section}. 
In general, the dynamics of a manipulator model can be expressed as:
\begin{equation}
     \mathbf{\ddot{q}} = \mathbf{M(\mathbf{q})}^{-1} (\boldsymbol{\tau}_u-\mathbf{C}(\mathbf{q},\mathbf{\dot{q}})\mathbf{\dot{q}} - \mathbf{G}(\mathbf{q})- \mathbf{J}^T\mathbf{W}_u)
    \label{eq:manip_model}
\end{equation}
where $\mathbf{q}$ is the joint state vector, $\mathbf{M(\mathbf{q})}$ is the joint space mass matrix, $\mathbf{C}(\mathbf{q},\mathbf{\dot{q}})$ is the Coriolis term, $\mathbf{G}(\mathbf{q})$ is the gravity term, $\boldsymbol{\tau}_u$ is the torque applied at joints, $\mathbf{J}$ is the end-effector Jacobian and $\mathbf{W}_u$ is the external Cartesian wrench at the end-effector.\par

\begin{remark}
In this study, manipulator and tool dynamics are modeled separately. The underlying motivation is driven by our experimental design procedure and is due to the fact that the manipulator model inertial parameters are not very well-identified.
\end{remark}

We model the application tool as a rigid body with a contact point and an wrench input $\v W_u$ as shown in Figure \ref{fig:modulation_setup}. 
This is independent from the manipulator model in the previous section. The Cartesian position of the contact point is defined as $\mathbf{x}_e = [x_e \hspace{4pt} y_e \hspace{4pt} z_e]_{\{world\}}$ in the ${\{world\}}$ frame. For brevity we omit the reference frame notation.
The tool dynamics can be represented as:
\begin{align}
     \begin{bmatrix}\mathbf{\ddot{x}}_c \\
     \boldsymbol{\dot{\omega}}_c
     \end{bmatrix} &= \mathbf{H}^{-1} \begin{bmatrix} \v W_u + m\mathbf{g} - \mathbf{F}_e \\ \mathbf{r}_{cb} \times \v W_u  + \mathbf{r}_{ce} \times \mathbf{F}_e  \end{bmatrix}
\end{align}
where $\mathbf{H} = {\rm diag}[m,\ldots,I_{xx},I_{yy},I_{zz}] \in \mathbb{R}^{6\times6}$, $m$ is the mass of the  \vspace{4pt} tool and $I_{xx, yy, zz}$ are the moments of inertia around center of mass. $\mathbf{\ddot{x}}_c$, $\boldsymbol{\vg \dot{{\omega}}}_c \in \mathbb{R}^3$  are the linear and angular centroidal dynamics of the tool while $\mathbf{F}_e \in \mathbb{R}^3$ is the force vector at the tool contact point.
\vspace{-2pt}
\begin{align*}
\boldsymbol{\omega}_e &=  \boldsymbol{\omega}_c,\quad
\boldsymbol{\dot{\omega}}_e =  \boldsymbol{\dot{\omega}}_c \\
[\ddot{\psi}_e \hspace{4pt} \ddot{\vartheta}_e \hspace{4pt} \ddot{\varphi}_e]^T &= \mathbf{T}^{-1}_e [\boldsymbol{\dot{\omega}}_e- \mathbf{\dot{T}}_e [\dot{\psi}_e \hspace{4pt} \dot{\vartheta}_e \hspace{4pt} \dot{\varphi}_e]^T]  
\end{align*}
where $\boldsymbol{\omega}_{c,e} = \mathbf{T} [\dot{\psi} \hspace{4pt} \dot{\vartheta} \hspace{4pt} \dot{\varphi}]^T$ is the angular rate of the tool, $\boldsymbol{\dot{\omega}}_{c,e} = \mathbf{T} [\ddot{\psi} \hspace{4pt} \ddot{\vartheta} \hspace{4pt} \ddot{\varphi}]^T + \mathbf{\dot{T}} [\dot{\psi} \hspace{4pt} \dot{\vartheta} \hspace{4pt} \dot{\varphi}]^T$, $\mathbf{T}$ is the corresponding mapping from the Euler rate\footnote{see Euler error section (chapter 2) of \cite{B.Siciliano1999RobotControl}} \cite{SicilianoB} to the angular rate. \par
The Cartesian path to track is defined at the tooltip. To obtain the tooltip dynamics, we project the centroidal dynamics to the tooltip along the rigid body by:
\begin{align*}
    \mathbf{\dot{x}}_e &= \mathbf{\dot{x}}_c + \boldsymbol{\omega}_c\times\mathbf{r}_{ce} \\
    \mathbf{\ddot{x}}_e &= \mathbf{\ddot{x}}_c + \boldsymbol{\dot{\omega}}_c\times\mathbf{r}_{ce} + \boldsymbol{\omega}_c\times(\boldsymbol{\omega}_c\times\mathbf{r}_{ce})
\end{align*}
\subsection{Contact model}\label{sec:contact model section}
Let the moving direction (unit vector) of the tool be $\mathbf{n}_v$ and the direction orthogonal to the moving direction as $\mathbf{n}_{\perp}$, where $\mathbf{n}_{\perp} \times \mathbf{n}_v = \mathbf{N}[i]$. $\mathbf{N}[i]$ is the surface average normal vector at $i^{th}$ time-step (as precept by the force-torque sensor). In this study, we use $\mathbf{N} = \begin{bmatrix}0 & 0 & 1 \end{bmatrix}^T_{\{world\}}$, when contact surface is horizontal.\footnote{In reality $\mathbf{N}[i]$ varies with the surface deformation and inclinations. It can estimated through an external force-torque sensor attached to the end-effector. For brevity, we represent $\mathbf{N}[i] \rightarrow \mathbf{N}$}. Then, we combine Eqs.~(\ref{eq:quasi-static-dir}, \ref{eq:friction_dyn}, \ref{eq:friction_dyn_orthogonal}) to obtain dynamics of the contact force vector $\mathbf{F}_e$.
\begin{multline}\label{contact model}
    \dot{\mathbf{F}}_e = \Big(\left(6E^{2}RF_z\right)^{\frac{1}{3}}\dot{d}\Big)\mathbf{n}_z \\ + \Big(\mu \dot{F_z}+\frac{3 \mu (2\nu-1)}{10R}\left(\dot{F_z}d+F_z\dot{d}\right)\Big)\mathbf{n}_v + \frac{2m v\dot{v}}{R_c}\mathbf{n}_{\perp}
\end{multline}

where $d$ is the deformation at central point of contact circle and calculated from Eq.~(\ref{eq:quasi_static_model}), and $\dot{d} = \v N^T\dot{\v x}_e=\dot{z}_e$. $F_z$ is the vertical force (the surface normal direction) applied on the surface by the manipulator and $v = \| \dot{\v x}_e\|_2$ is the moving velocity of the tool contact point.

\subsection{Trajectory optimization formulation}
\label{sec:overall dynamic model}
The state for our trajectory optimization is represented as:
the end-effector state $\mathbf{x}_E = [x_e \hspace{4pt} y_e \hspace{4pt} z_e \hspace{4pt} \psi_e \hspace{4pt} \vartheta_e \hspace{4pt} \varphi_e]$, the 7-DOF manipulator joint state $\mathbf{x}_M \in \mathbb{R}^7$, 
 $\mathbf{x} = [\mathbf{x}_M \hspace{4pt} \dot{\mathbf{x}}_M \hspace{4pt} \mathbf{x}_E \hspace{4pt} \dot{\mathbf{x}}_E \hspace{4pt} \mathbf{F}_e]^T, 
 \mathbf{u} = [\v W_u \hspace{4pt} \boldsymbol{\tau}_u]$.
Orientation is represented by Euler angles.

The overall optimization problem is formulated as: 
\begin{subequations}\label{eq:prime_objective}
\begin{align}\label{eq:local_cost}
    &\underset{\vg \phi}{\text{min}} \ &&\sum_{i = 0}^{N} \hspace{2pt}  \overbrace{\delta\mathbf{F}[i]^T\mathbf{Q}_F\hspace{2pt}\delta\mathbf{F}[i]}^{\text{force tracking}}
    + \mathbf{u}[i]^T\mathbf{R}\mathbf{u}[i] \nonumber \\  &&&+ \hspace{2pt} \underbrace{W_p\big\| \mathcal{FK}(\mathbf{x}_M^c[i])-\mathbf{x}_e^d[i]\big\|_2}_{\text{pose tracking}} \\\label{eq:general_dynamics}
    \nonumber \vspace{2pt} \\ 
    &\text{s.t.} \quad 
    &&\mathbf{x}[i+1] = \mathcal{F}(\mathbf{x}[i], \mathbf{u}[i]), \forall i = 1, \ldots, N-1 \\
    &&&
    \mathbf{x}[0] = \mathbf{x}_0 \\\label{eq:joint_limit}
    &&&\underline{\mathbf{x}}_M \leq \mathbf{x}_M \leq \bar{\mathbf{x}}_M \\\label{eq:wrench_limit}
    &&&\underline{\v W}_u \leq \v W_u \leq \bar{\v W}_u  \\&&&\frac{m v^2}{R_c} \le \mu \mathbf{N}^T\mathbf{F}_e\mathbf{N} + \mathcal{G}(\mathbf{x}_e) \label{eq:centripetal_force}
\end{align}
\end{subequations}
where $\delta\mathbf{F}[i] = (\v F_c[i] - \v F_{d}[i])$ is the current force state error with respect to reference $\v F_d[i]$,  $\mathbf{Q}_F \in \mathbb{R}^{3\times 3}$ and $\mathbf{R} \in \mathbb{R}^{m\times m}$ are the state and control weighting matrices, $\mathcal{FK}(\cdot) \in SE(3)$ is the forward kinematics function for the manipulator and $\mathbf{x}_e^d[i] \in SE(3)$ is the desired Cartesian pose. $W_p$ is a scalar weighting coefficient.   $\mathbf{x}_M^c[i]$, $\mathbf{x}_M^d[i] \in \mathbb{R}^7$ are the current and desired joint vectors. For simplicity, we use $\vg \phi=(\v x[0,\ldots,N],\v u[0,\ldots,N-1])$ to represent the sequence of state-control pairs. $v = \|\dot{\mathbf{x}}_e\|_2$ and $R_c$ is the curvature (see Figure \ref{fig:modulation_setup}) of the radius of the path tracked and $\mathcal{G}$ is a general term to compensate for added deformation (rolling) friction. Eq. (\ref{eq:centripetal_force}) represents the frictional constraint in the orthogonal direction to the moving direction to maintain the motion.

\subsection{Manipulator wrench control}
\label{manipulator control section}
The manipulator used in our experiment was a 7-DOF KUKA LBR R820. Inertial parameters for this manipulator were not accurately identified to the precision of generating small external wrenches using torque control. As an alternative, KUKA FRI (Fast Research Interface \cite{KukaManual}) allows users to accurately command external wrenches at the end effector. It uses the proprietary internal dynamic model of the manipulator to generate commanded external wrenches. This allows us to define a desired Cartesian impedance for the manipulator. The commanded input applies a wrench to the base of the application tool. The manipulator control law is expressed with combined feed-forward and feedback components as:
\begin{equation}
    \mathbf{\bar{W}}_u = \mathbf{W}_u + \mathbf{K}\hspace{2pt}(\mathbf{x}_{e,c}-\mathbf{x}_{e,d})
    \label{eq:control_law}
\end{equation}
where $\mathbf{K} \in \mathbb{R}^{3\times3}$ stands for the user specified proportional gain, i.e., stiffness in cartesian space. $\mathbf{x}_{e,c}-\mathbf{x}_{e,d} \in \mathbb{R}^3$ is the distance between the current pose and the desired pose of end-effector (with the tool part).\footnote{Note that, $\mathbf{x}_e^d$ and $\mathbf{x}_{e,d}$ in this paper represent different Cartesian states.} In our optimal control problem in Eq.~(\ref{eq:prime_objective}), we take the external wrench ($\mathbf{W}_u$) 
as a part of the control input. The control input was implemented on the manipulator with a 
closed-loop position control to mitigate error compounding as governed by the Eq.~(\ref{eq:control_law}).

\section{Constrained Trajectory Optimization with Contact Dynamics}
Given the manipulator and tool models in Sec.~\ref{manipulator and tool models section} and the proposed contact dynamics in Sec.~\ref{sec:contact model section}, Differential Dynamic Programming (DDP) is used in this section to generate desired joint and Cartesian motion as well as force profiles obeying dynamic constraints.

DDP is well received for effectively solving unconstrained trajectory optimization. It represents an indirect method which only optimizes control inputs, and the dynamics constraint is implicitly satisfied during the forward trajectory rollout. Given an initial guess of control inputs, an updated state trajectory is generated by forward propagating the differential equation of rigid-body dynamics. Then a quadratic approximation is constructed for the cost function and dynamics around the current trajectory so that a Riccati recursion can be used to derive the optimal feedback control law. By iteratively updating the state and control trajectories, the optimization will converge to an optimal solution. 

One limitation of DDP stems from its difficulty in addressing constraints. Since our contact model enforces state, control, frictional constraints, it is desired to incorporate these contact constraints along with the state and control constraints. The work in \cite{ZhouAcceleratedDynamics} proposed an iterative solve process based on Alternating Direction Method of Multipliers (ADMM) to address a variety of constraints. In this work, we will employ this ADMM to handle the constraints, in particular those involved in contact dynamics.

The ADMM algorithm decomposes a large-scale, holistic optimization problem into sub-problems and solves each sub-problem iteratively. In each iteration, the primal and dual variables are updated sequentially. Under mild conditions, both primal and dual variables converge to the optimal solutions. More details about ADMM formulations are referred to \cite{boyd2011distributed}. To apply this algorithm for our soft-contact manipulation problem, we transcribe the original optimal control problem (\ref{eq:prime_objective}) into the following form:
\begin{align}\nonumber
     \underset{\vg \phi, \hat{\vg \phi}}{\text{min}} \ &\sum_{i=0}^{N} \mathcal{C}(\v x[i], \v u[i])
        + I_{\mathcal{D}}(\v x[i], \v u[i]) \\&+ I_{\mathcal{J},\mathcal{U},\mathcal{F}}(\hat{\v x}_M[i], \hat{\v u}[i],\hat{\vg \lambda}[i])\\\nonumber
        \text{s.t.} \ &\v x_M = \hat{\v x}_M, \ \v u=\hat{\v u},\ \vg \lambda = \hat{\vg \lambda}
\end{align}
where we define $\mathcal{C}(\v x, \v u)$\footnote{Due to space limit, we ignore the summation notation and time-step subscripts in ADMM updates. For example, $\mathcal{C}(\v x, \v u)=\sum_{i=0}^{N} \mathcal{C}(\v x[i], \v u[i])$.} as the local cost function corresponding to Eq.~(\ref{eq:local_cost}) and $\vg \lambda=({\dot{\v x}_e}^T,{\v F_e}^T)^T$. We use $\hat{\vg \phi}=(\hat{\v x}_M[0,\ldots,N],\hat{\v u}[0,\ldots,N-1],\hat{\vg \lambda}[0,\ldots,N])$\footnote{The decision variables $\v x_M$ and $\vg \lambda$ are subsets of the full state $\v x$} to express the concatenated states and controls that are required to be projected. The admissible set $\mathcal{D}$ represents the generalized dynamics constraint (\ref{eq:general_dynamics}), where $\mathcal{D}=\{(\v x,\v u) \ | \ \v x[0]=\v x_{\rm init}, \v x[i+1]=\mathcal{F}(\v x[i],\v u[i]), i=0,1,\ldots,N-1  \}$. The closed and convex sets $\mathcal{J}$, $\mathcal{U}$ and $\mathcal{F}$ stand for joint limit (\ref{eq:joint_limit}), control limit (\ref{eq:wrench_limit}) and contact constraint (\ref{eq:centripetal_force}), respectively. In general, an indicator function regarding a closed convex set $B$ is defined as
\begin{equation}
    I_B(\v x,\v u)=\begin{cases}
    0, \ (\v x,\v u)\in B \\
    +\infty, \ \text{otherwise}
    \end{cases}
\end{equation}

Then for each ADMM iteration $k$, the updating sequence with scaled dual variables is
\begin{subequations}\label{eq:updates}
    \begin{align}
       \label{eq:primal-a} \nonumber
        \vg \phi^{k+1}=&\underset{\vg \phi}{\arg\min} \ \mathcal{C}(\v x, \v u)
        + I_{\mathcal{D}}(\v x, \v u) + \frac{\rho_j}{2}\|\v x_M-\hat{\v x}_M^{k}+\v v_j^k\|_2^2\\
         & + \frac{\rho_u}{2}\|\v u-\hat{\v u}^{k}+\v v_u^k\|_2^2 + \frac{\rho_f}{2}\|\vg \lambda-\hat{\vg \lambda}^{k}+\v v_f^k\|_2^2\\\label{eq:primal-b} \nonumber
        \hat{\vg \phi}^{k+1}=&\underset{\hat{\vg \phi}}{\arg\min} \ I_{\mathcal{J},\mathcal{U},\mathcal{F}}(\hat{\v x}_M, \hat{\v u},\hat{\vg \lambda})+\frac{\rho_j}{2}\|\v x_M^{k+1}-\hat{\v x}_M+\v v_j^k\|_2^2  \\
         &+ \frac{\rho_u}{2}\|\v u^{k+1}-\hat{\v u}+\v v_u^k\|_2^2 + \frac{\rho_f}{2}\|\vg \lambda^{k+1}-\hat{\vg \lambda}+\v v_f^k\|_2^2\\
        &\hspace{0.3in}\v v_j^{k+1}=\v v_j^{k}+\v x_M^{k+1}-\hat{\v x}_M^{k+1}\\&\hspace{0.3in}\v v_u^{k+1}=\v v_u^{k}+\v u^{k+1}-\hat{\v u}^{k+1}\\&\hspace{0.3in}
        \v v_f^{k+1}=\v v_f^{k}+\vg \lambda^{k+1}-\hat{\vg \lambda}^{k+1}
    \end{align}
\end{subequations}
where $\vg \phi$ and $\hat{\vg \phi}$ are primal variables. $\v v_j$, $\v v_u$, $\v v_f$ are dual variables related to joint limits, control limits and contact constraints. $\rho_j$, $\rho_u$ $\rho_f$ are step-size parameters corresponding to each constraint. This scaled form integrates linear and quadratic terms in augmented Lagrangian which is compact and easier to work with.

Note that for Eq.~(\ref{eq:primal-a}), we use DDP to solve it and $I_{\mathcal{D}}$ is always zero since the state trajectory is always dynamically feasible by performing the forward pass. For Eq.~(\ref{eq:primal-b}), this minimization problem reduces to a projection operator on convex sets $\mathcal{J}$, $\mathcal{U}$, and $\mathcal{F}$
\begin{equation}
    \begin{aligned}\nonumber
       \hat{\vg \phi}^{k+1}=&\underset{\hat{\vg \phi}\in C}{\arg\min} \ 
         \frac{\rho_j}{2}\|\v x_M^{k+1}-\hat{\v x}_M+\v v_j^k\|_2^2 \\&+ \frac{\rho_u}{2}\|\v u^{k+1}-\hat{\v u}+\v v_u^k\|_2^2 + \frac{\rho_f}{2}\|\vg \lambda^{k+1}-\hat{\vg \lambda}+\v v_f^k\|_2^2\\
         C=&\{(\hat{\v x}_M,\hat{\v u},\hat{\vg \lambda})|\hat{\v x}_M \in \mathcal{J}, \hat{\v u} \in \mathcal{U},\hat{\vg \lambda} \in  \mathcal{F}\}
    \end{aligned}
\end{equation}
We use a saturation function to efficiently project the infeasible values onto the boundaries defined by different constraints. The whole process of our ADMM algorithm is shown in Algorithm 1. We initialize $\hat{\vg \phi}$ and dual variables $\v v$ as zero. The initial trajectory of $\vg \phi$ is generated by running the forward pass with an initial guess of controls. In each ADMM iteration, the controls from last ADMM iteration will warm start the current DDP solver, which makes the DDP solver converges faster within around ten iterations in each ADMM iteration. Then the trajectories are solved iteratively until a stopping criterion with regard to primal residuals (see \cite{boyd2011distributed}, Sec.~3.3) is satisfied.
\begin{algorithm}\label{pseudo:DDO-ADMM}
  \caption{ADMM trajectory optimization}
  \begin{algorithmic}[1]
   \STATE $\vg \phi \gets \vg \phi^{0},\hat{\vg \phi} \gets \hat{\vg \phi}^{0}$
   
   \STATE $\v v_j \gets \v v_j^0,\v v_u \gets \v v_u^0,\v v_f \gets \v v_f^0$
   
    \REPEAT
    \STATE $\vg \phi \gets \text{DDP}\;(\vg \phi,\hat{\v x} - \v v_{j},\hat{\v u}-\v v_{u},\hat{\vg \lambda}-\v v_{f})$
    \STATE $\hat{\vg \phi} \gets \text{Projection}\;(\v x+\v v_j,\v u+\v v_u,\vg \lambda+\v v_f)$
    \STATE $\v v_j \gets \v v_j + \v x - \hat{\v x}$
    \STATE $\v v_u \gets \v v_u + \v u - \hat{\v u}$
    \STATE $\v v_f \gets \v v_f + \vg \lambda - \hat{\vg \lambda}$
    \UNTIL{$\rm{stopping\ criterion\ is\ satisfied}$}
    \RETURN{$\vg \phi$}
  \end{algorithmic}
\end{algorithm}

\section{Experimental Validation}
\label{experimental validation}
\begin{figure}[t] 
\centering
\includegraphics[width=0.7\linewidth,height=4cm]{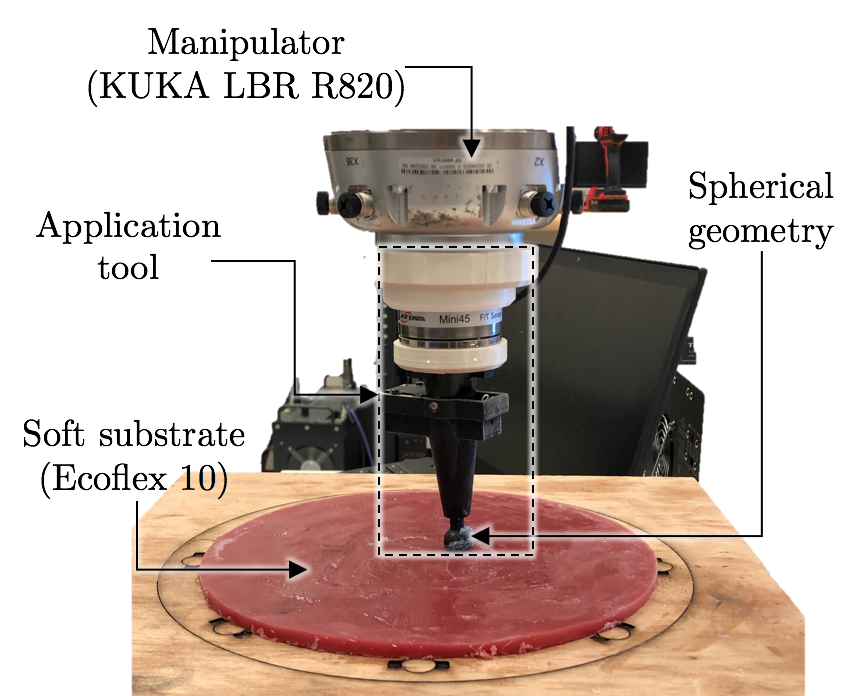}
\caption{Experimental setup. The application tool with a spherical geometry in contact with the soft material.}
\label{fig:experimental setup}
\vspace{-10pt}
\end{figure}
\subsection{Identification of material properties}
To experimentally validate the proposed soft contact model, parameters related to contact body material need to be identified, e.g., frictional coefficient and Young Modulus. 
\begin{figure}[t] 
\centering
\includegraphics[width=1.0\linewidth,height=4cm]{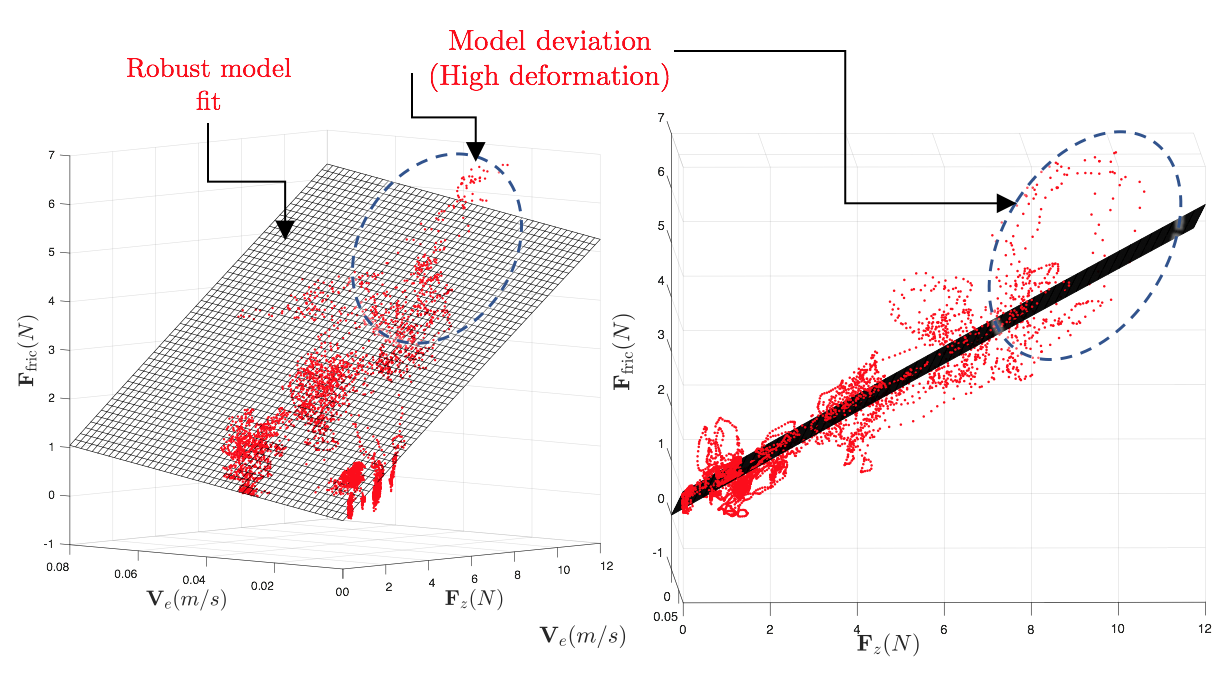}
\caption{Friction model validation and identification. $\mu = 0.4512, k_d = 13.1315$ Ns/m, $\text{R-squared} = 0.9103$.}
\label{fig:friction_id}
\vspace{-10pt}
\end{figure}

\begin{figure*}[t] 
\centering
\includegraphics[width=1.0\textwidth,height=10cm]{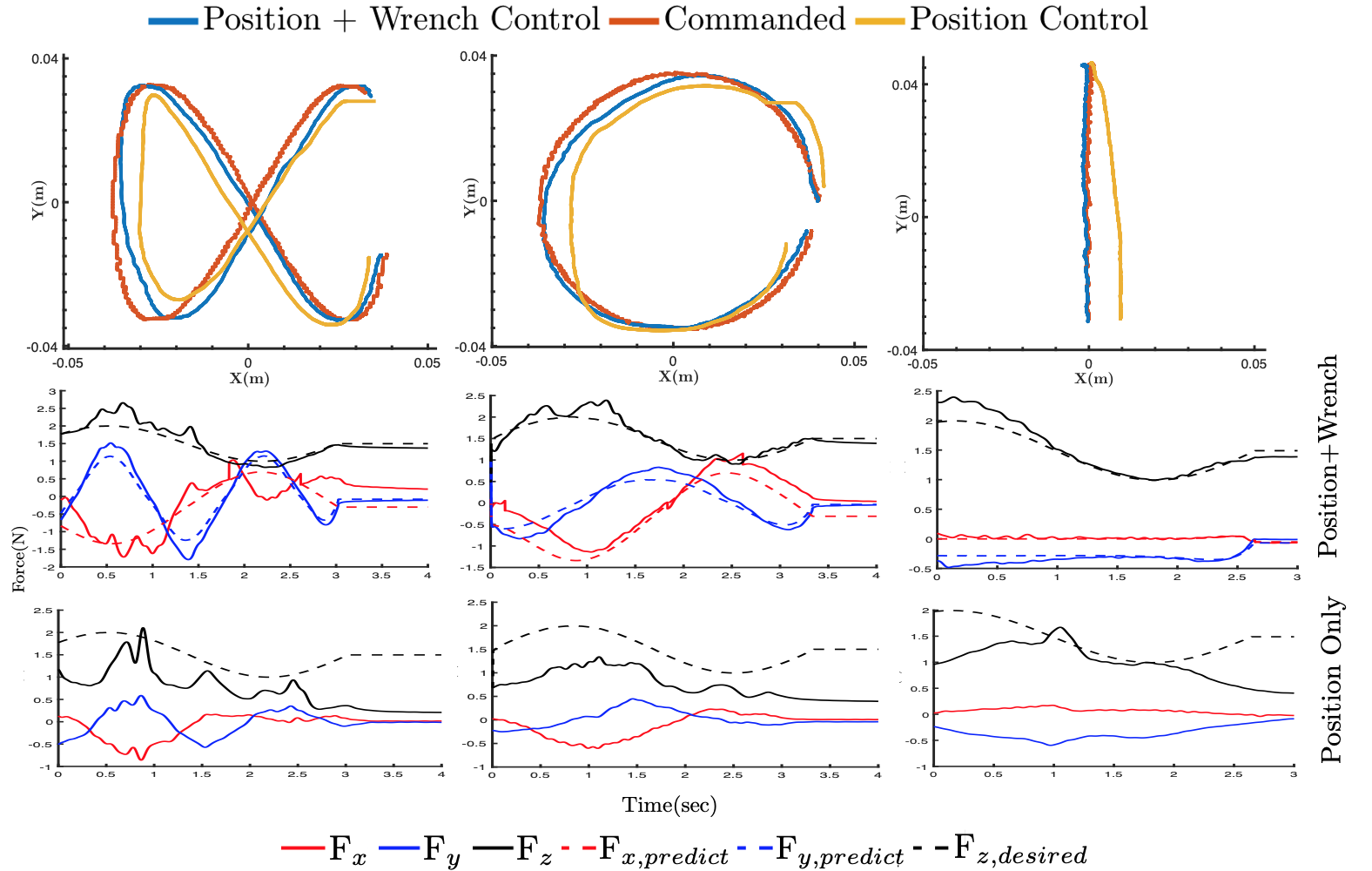}
\caption{\textit{Top row:} Comparison of the Cartesian paths tracked by the end-effector with wrench control input generated by ADMM and with simple position control.
\textit{Middle Row:} Comparison of the normal and frictional forces predicted by ADMM and the ground truth force data.
\textit{Bottom Row:} Ground truth force data recorded in simple position control. $\v F_d = [0 \quad 0 \quad  F_{z,desired}]^T$ is the desired tracking force.}
\label{fig:results_open_loop}
\vspace{-10pt}
\end{figure*}

\par
The Young Modulus was estimated through performing cyclic linear probing on the surface of the material with the same end-effector point geometry of a sphere (was tested on a material testing platform INSTRON$^\copyright$). The Young Modulus was estimated through a non-linear least square estimator by using Eq.~(\ref{eq:quasi_static_model}).  
The friction coefficient was estimated by performing motions along the material surface while recording the force/torque data through an ATI mini45 sensor attached to the end-effector as shown in Figure \ref{fig:experimental setup}. Eq.~(\ref{eq:friction_model}) was used to estimate the frictional coefficients. Frictional force magnitude in the moving direction $\v F_{\rm fric}$, velocity magnitude $\v V_{e}$, and normal contact force $F_{z}$ were calculated from the collected data (see Fig.~\ref{fig:friction_id}) and a three dimensional robust least square approximation was done (with a logistic distance function in MATLAB$^\copyright$). Robust least square fitting was used to mitigate the sensitivity to the model deviation as the deformation increases.
Identified Young Modulus and friction coefficient were incorporated into the optimization in Eq.~(\ref{eq:prime_objective}). Desired states to track were the desired end-effector position ($x_e, y_e,z_e$) and the desired contact force $F_z$. Optimal trajectories and inputs (Cartesian wrench $\v W_u$) were generated through the optimization problem of Eq.~(\ref{eq:prime_objective}). Constraints were satisfied within $~10^{-2}$ residual violations in both primal and dual stopping criteria. Desired contact force ($\v F_d$) was maintained within the valid range of the friction model. \par

The material surface properties were identified to validate the derived contact model and further use it in the trajectory optimization to generate optimal open-loop trajectories. Model fitting to the friction data is shown in Figure \ref{fig:friction_id}. Data was fit with an resulting R-squared value of $0.9103$ and frictional coefficient ($\mu = 0.4512$) and damping coefficient ($k_d=13.1315$ Ns/m) were identified. \par
It was observed that with the increase of normal force on the surface, the effects of deformation dominated the frictional force data obtained through model fitting. This phenomenon was due to the increased rolling friction and material-specific artifacts, e.g., non-uniformity in frictional coefficient and stress distribution. Moreover, the presence of
fluids or any micro-particular particles will increase the non-uniformity.
\subsection{Motion along a desired path with desired contact force}
Open-loop trajectories generated from our trajectory optimization were input to the manipulator (i.e., Cartesian wrench, $\v W_u$, as described in section \ref{manipulator control section}). We generated Cartesian trajectories along a line, a circle, and a number eight shape, respectively. Figure \ref{fig:results_open_loop} shows the experimental results where the top row depicts motion tracking while the middle and bottom rows show the predicted and actual values of $F_x, F_y, F_z$ in the ``Position+Wrench Control`` and ``Position Control`` modes, respectively. The optimization generates desired Cartesian trajectories ($\mathbf{x}_e^d$) and force profile ($\v F_d$) We compared the results of running the control input ($\v W_u$) generated from the optimizer along with a simple position control where the feed-forward wrench component in Eq.~(\ref{eq:control_law}) is dropped. Since the control was in wrench mode, motion and forces are coupled. Any mismatch in contact forces (e.g., those due to friction and deformation) would directly affect the motion and vice versa. It was observed that the control input solved via trajectory optimization was able to track the motion ($\mathbf{x}_e^d \in SE(3)$) and force profile ($\v F_d$) significantly better than simple position control.  However, the tracking performance is not superior but adequate due to the unmodelled dynamics, stiction, and non-uniformity of the soft material.

\section{Conclusion and Future Work}
\label{sec:future work}
For automation tasks involving biological or soft tissue contact, it is paramount to design soft contact interaction models where controllers can be designed to guarantee safe tracking performance.
Identifying valid contact models and incorporating them in robot motion generation is essential to safely and efficiently automate mundane contact-rich tasks. Some examples include incisions, motion and force compensation on soft bodies under external disturbances such as breathing or heartbeat.
This study presents a coherent framework to perform simultaneous motion and force modulation on compliant surfaces. Potential applications of this work include contact manipulation in soft tissues or safety-critical environments. Moreover, this framework can be generalized to any applicable contact model with identified parameters and the manipulator model. 
\par In summary, we derived a soft contact dynamical model and incorporated it into a trajectory optimizer capable of handling state, control, and contact constraints. Trajectories were solved using the ADMM trajectory optimizer, which included two sub-optimization problems; namely, DDP and projection blocks. Obtained trajectories were experimentally validated on a soft surface (EcoFlex$^\copyright$) with the aid of a robot manipulator with an attached spherical shaped tooltip. Surface material properties were estimated and used in generating optimal trajectories. Ground truth forces were obtained using a force-torque sensor (ATI mini45) and compared against the predicted results. Results show accurate tracking of the forces and desired positions as predicted by the derived dynamic contact model.
\par 
In the future, we intend to split the optimization cost to multiple blocks ($N_{\rm block}>2$) and implement the trajectory optimization in a real-time model predictive control (MPC) fashion. Moreover, a low-level adaptive controller will be used to handle material non-uniformity and uncertainty caused by environmental disturbances.




\section*{APPENDIX}
\label{sec:appendix}
\noindent
In this Appendix, we provide the details on several stress distributions. First, the normal Stress Distribution $\sigma_z$:
\begin{equation}
 \frac{\sigma_z}{p_m}=-\frac{3}{2}\left(1-\frac{r^2}{a^2}\right)^{\frac{1}{2}}\quad (r\leq a)
\end{equation}
\\
Radical Stress Distribution $\sigma_r$:
\begin{equation*}
     \frac{\sigma_r}{p_m}=\frac{2\nu-1}{2}\frac{a^2}{r^2}\left[1-\left(1-\frac{r^2}{a^2}\right)\right]-3\nu\left(1-\frac{r^2}{a^2}\right)^{\frac{1}{2}}\hspace{1pt} (r\leq a)
\end{equation*}
Hoop Stress Distribution $\sigma_\theta$:
\begin{equation*}
     \frac{\sigma_\theta}{p_m}=\frac{1-2\nu}{2}\frac{a^2}{r^2}\left[1-\left(1-\frac{r^2}{a^2}\right)\right]-\frac{3}{2}\left(1-\frac{r^2}{a^2}\right)^{\frac{1}{2}}\hspace{1pt} (r\leq a)
\end{equation*}
where $p_m=F/(\pi a^2)$ is the average stress applied in contact part by manipulation and $a=\sqrt{Rd}$ is the radius of contact area (refer to figure \ref{fig:contact_ball}). The transformation matrix $T$ is 
\begin{equation*}
   T= {\left[ \begin{array}{ccc}
c\theta & s\theta & 0\\
-s\theta &c\theta & 0\\
0 & 0 & 1
\end{array} 
\right ]}
\end{equation*}
\\


\bibliographystyle{ieeetr}
\bibliography{refs_LW}

\end{document}